\documentclass[review,12pt]{elsarticle}

\usepackage{amsmath,amssymb}
\usepackage{bbm}

\usepackage{changepage}

\usepackage{textcomp,marvosym}

\usepackage{nameref,hyperref}

\usepackage{algorithmic}
\usepackage[ruled,vlined]{algorithm2e}
\usepackage{tablefootnote}
\usepackage{gensymb}

\journal{Pattern Recognition}

\begin{document}

\begin{frontmatter}



\title{Efficient High-Resolution Template Matching with Vector Quantized Nearest Neighbour Fields}


\author[inst1]{Ankit Gupta\corref{cor1}}
\author[inst1]{Ida-Maria Sintorn}
\affiliation[inst1]{organization={Department of Information Technology, Uppsala University},
            city={Uppsala},
            postcode={752 37}, 
            country={Sweden}}

\cortext[cor1]{Corresponding Author.\newline E-mail addresses: ankit.gupta@it.uu.se, ida.sintorn@it.uu.se
}
\begin{abstract}

Template matching is a fundamental problem in computer vision with applications in fields including object detection, image registration, object tracking. Current methods rely on nearest-neighbour (NN) matching, where the query feature space is converted to NN space by representing each query pixel with its NN in the template. NN-based methods have been shown to perform better in occlusions, appearance changes, and non-rigid transformations; however, they scale poorly with high-resolution data and high feature dimensions. We present an NN-based method which efficiently reduces the NN computations and introduces filtering in the NN fields (NNFs). A vector quantization step is introduced before the NN calculation to represent the template with $k$ features, and the filter response over the NNFs is used to compare the template and query distributions over the features. We show that state-of-the-art performance is achieved in low-resolution data, and our method outperforms previous methods at higher resolution.


\end{abstract}




\end{frontmatter}

\section{Introduction}
Template matching refers to locating a small template image, $T$, in a larger image, $I$. It is a fundamental problem in computer vision with applications in various fields such as object detection \citep{lucas2021locating,thomas2020multi}, object tracking \citep{wang2021automatic,xu2021accurate}, document information identification \citep{sun2019template}, and image registration \citep{ye2019fast}. It also has a central role in deep learning due to the increasing demand for annotated data. Template matching has, for example, been used as a tool for human-in-the-loop data annotation frameworks \citep{thomas2020multi,gupta2022simsearch} as it offers fast detection at a relatively low cost of labour and resources. Here, we refer to human-in-the-loop data annotation as object detection on an image set where one or multiple instances of different classes can be present. Template matching allows users to find similar objects quickly without expensive classifier training.     

Traditional template matching approaches such as sum-of-squared distance (SSD), sum-of-absolute distance (SAD), and normalized cross-correlation (NCC) are very efficient. However, they consider every pixel pair in the sliding subwindow of query image $I$ and $T$, making them vulnerable to occlusion and transformations. More recently, nearest neighbour field (NNF) based matching approaches \citep{talker2018efficient,lan2021gad} have been suggested and shown to overcome these shortcomings, making them state-of-the-art in the field. NNFs constitute a general and non-parametric framework for generating correspondences between the sub-regions of images and have been successfully used for motion-tracking in videos \citep{zhou2015abrupt,ben2015approximate}, optical flow algorithms \citep{chen2013large}, and structural image editing \citep{barnes2009patchmatch}. NNF-based template matching approaches rely only on the subset of ``good" matches between the template and the query subwindow. This makes them more robust against complex non-rigid transformations, occlusions, and background clutter. They achieve this by matching between two point sets, namely, the template point set and the query point set. A similarity measure is defined between the sets based on the matching statistics, for example, bi-directional matches \citep{dekel2015best} or unique matches \citep{talmi2017template}.

NNF-based approaches represent the points in the query image with the nearest neighbour of the points in the template image. This formulation reduces the pixel representation space in relation to the template. For example, in an 8-bit RGB image, the possible pixel values can be from a set size of $|255|^3$, but representing the image with the NNF of a template size $|w\times h|$ will reduce the representation set to $wh$ thus greatly lowering the set of possible values. It can be viewed as a form of vector quantization \citep{KHAN2005673} where the codebook is defined by all the template pixels, and the query image pixels are then quantized with respect to the codebook. The codebook is, however, still too large for practical implementations and contains redundant information, as the representation of nearby pixels is likely very similar.

An inadvertent effect of using too many points in the NNF representation is the difficulty in considering the deformation implied in the NNF in the similarity score. To model the deformation in the NNF, previous methods rely on a deformation measure for each pixel in the query subwindow, penalising large relative distances between the template point location and its NN match location. This relies on the strong assumption of the similar orientation of the template and query subwindow, which makes these methods incapable of handling significant rotational or deformation changes. It also requires more computations as the relative distance has to be calculated for each point in the query sub-window. 

\begin{figure}[tb]
\centering
\includegraphics[width=1\columnwidth]{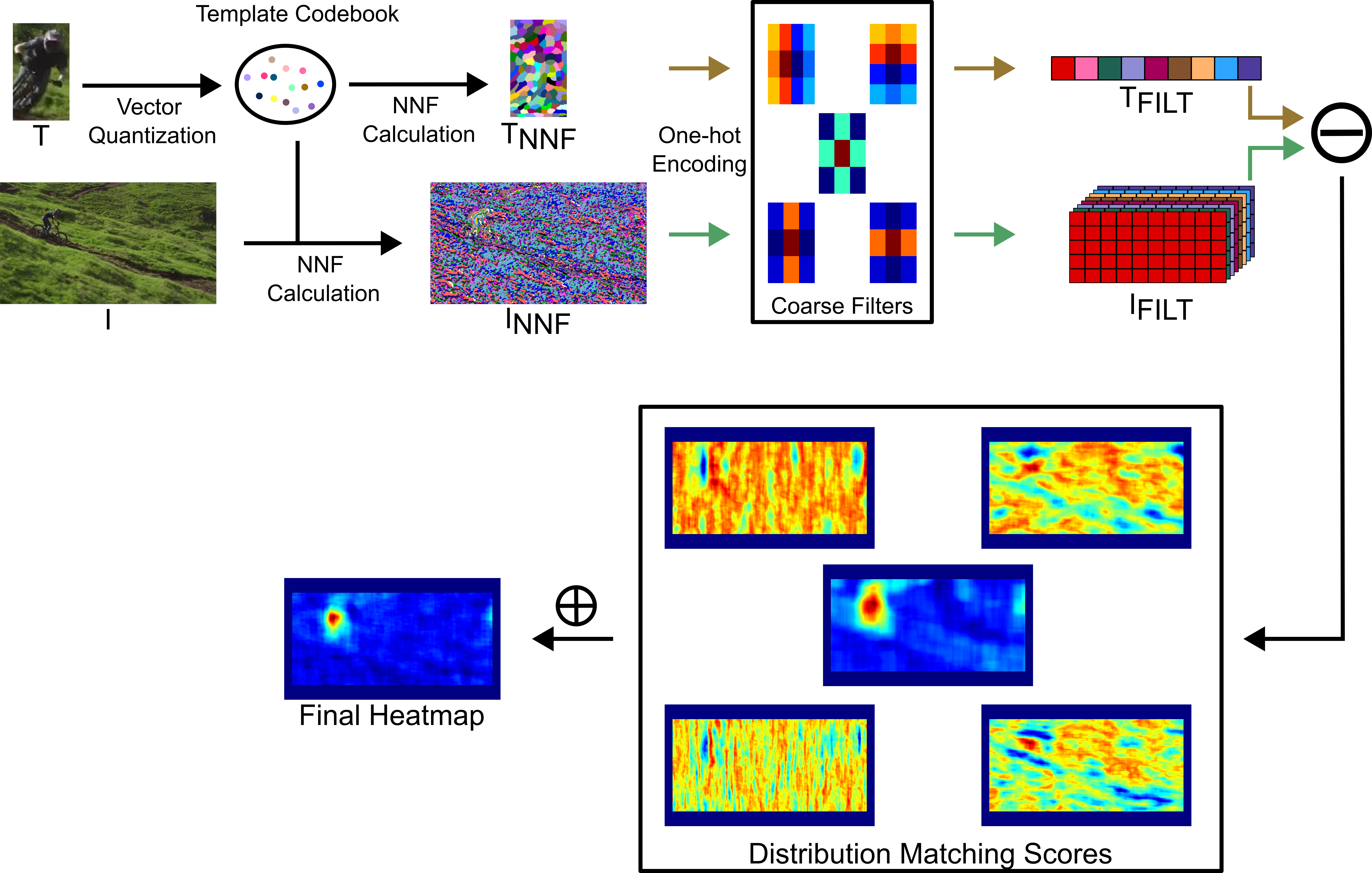}
\caption{The proposed template matching method. First, a $k$-sized vector-quantized codebook is constructed using the template points. The codebook is used to generate the NNF label image for the template and the query image. Multiple template representations are generated using coarse filters and compared with the filtered query distribution. The matching scores from different filters are combined to get the final heatmap.
}
\label{fig:intro_fig}
\end{figure}

In this work, we introduce a fast and flexible template-matching approach that reduces the NN calculations significantly. We reduce the template codebook size significantly to $k$ $(\ll wh)$ points by only considering major patterns in the template. This changes the computational complexity in the matching step to depend on only $k$ points instead of the number of template pixels, greatly reducing the NN computation complexity and NNF creation time. The difference in the codebook pattern distribution in the template and query subwindow is used as the dissimilarity measure for the matching. Since the codebook is small, simple coarse filters can be used to model the deformation in the NNF instead of considering pixel-wise distances.
Our major contributions are: 
\begin{itemize}
    \item We present a simple and fast NNF-based template matching method that greatly reduces computational costs in the NNF creation while maintaining performance.
    \item We introduce filtering in the NNF space to model the deformation using coarse filters and show that state-of-the-art performance can be achieved with simple Gaussian and Haar filters.
    \item We show that our method's quantitative performance and run-time scale better with the image resolution than previous approaches.
\end{itemize}

\section{Related Work}

Traditional methods in template matching, such as SSD, SAD, and NCC, work well in cases where only translational variance is present. They are very sensitive to deformations and non-rigid transformations. Classical algorithms that use SSD and SAD as the dissimilarity measure are comprehensively reviewed in \citep{ouyang2011performance}.

Different approaches have been proposed to model affine transformations between the template and query window \citep{korman2013fast,jia2016colour}. More recently, \citep{kat2018matching} proposed an approach based on template feature co-occurrence matrix statistics for the matching algorithm reaching state-of-the-art performance on the BBS datasets. In \citep{korman2018oatm}, a method is proposed that provides a quadratic improvement in the search complexity while also being robust to partial occlusion by formulating the template matching as a consensus set maximization problem, i.e., finding the transformation where the maximum number of pixels between the template and query window are co-visible. In \citep{yang2019robust}, a superpixel region binary descriptor (SRBD) is suggested to construct a multilevel semantic fusion vector used in the matching. First, the template is divided into superpixels by the KD-SLIC \citep{achanta2012slic} algorithm, whereafter, a region binary vector is constructed by describing the dominant superpixel orientation. The method is robust against large deformations, occlusions, noise and illumination variations with the computation complexity of $O(360\cdot K\cdot |I|)$ where $K$ is the number of superpixels. Template matching was done using a probabilistic inference known as ``explaining away" in \citep{spratling2020explaining} to consider the evidence for the template at each location in the query image. The similarity between the template and a query patch considers the pixel intensity similarity at the corresponding locations and the alternative explanations for the query intensity values represented by the same template at other locations. Different shape-texture training methods for CNNs were tried in \citep{gao2022shape} to find a feature space to improve template matching performance and combined with the previous approach \citep{spratling2020explaining} to make it more tolerant to appearance changes.

NNF-based methods have shown great promise due to their robustness against occlusion and deformations. In \citep{dekel2015best}, properties of the nearest-neighbour (NN) matches between the template and query image pixels were used to introduce the Best-Buddies Similarity (BBS) measure. BBS measures the similarity by counting the bidirectional NN matches between the template and query subwindow points, with more matches indicating a higher degree of similarity. This bidirectional matching focuses only on the relevant corresponding features, thus making it robust against deformations and occlusion, outperforming previous state-of-the-art. This approach is slow and hard to use in practice as the computational complexity for BBS calculation is $O(|I|\cdot|T|\cdot|I|)$ (where $|I|$ is the size of the query image and $|T|$ is the template size). In \citep{talmi2017template}, inspired by the idea of matching objects for texture synthesis using the patch diversity \citep{jamrivska2015lazyfluids}, diversity similarity (DIS) and deformable DIS (DDIS) measures were proposed that only rely on the NN matching in a single direction. This approach is faster than BBS; however, unique matches still have to be calculated over every query sub-window with the computational complexity of $O(|I|\cdot|T|)$, and it still remains time-consuming for larger image and template sizes. To reduce the computational complexity of the methods proposed above while retaining the performance, \citep{talker2018efficient} proposed image-based unpopularity (IWU) and deformable IWU (DIWU). They use the diversity within the query image NNF instead of the subwindow of NNF, reducing the complexity to $O(I)$. In \citep{lai2020fast}, a majority neighbour similarity and annulus projection transformation were used to provide a fast template matching method, which is also robust to different challenges. Recently, \citep{lan2021gad} proposed a global-aware diversity-based method, GAD, which combines the IWU and DIS scores to propose a parameter-free algorithm. A scale-adaptive NNF-based method was proposed in \citep{zhang2020scale}. It extends DIS through statistical analysis, which makes it robust to outliers. However, the method performs calculations over multiple scales, which is time-consuming.

\section{Method}
We define the input to our template matching method as a template $\tau$ of size $w\times h \times c$ and a query image $I$ of size $W \times H \times c$, where $c$ is the number of channels. The goal is to find the $w\times h \times c$ subwindow $q$ in $I$ most similar to $\tau$. To do this, a similarity score $S_{q_i}$ is calculated for each query subwindow ${q_i}$ using the sliding window procedure, and the subwindow with the highest score is our desired output. 

Our template-matching approach consists of three main steps: codebook and NNF creation, filtering in NNF, and similarity scoring, each described in detail below. 

\subsection{Codebook and NNF creation}
In previous methods, the NNF of the query image is generated by finding the nearest neighbour in the template for each pixel in the image (See Algorithm \ref{algo:nnf_codebook} in Supplementary Sec.~\ref{supp:nnf_create}). As mentioned in the Introduction, this can be reformulated as vector quantization of the query image, with a codebook of size $wh$ representing all the pixels in the template. This approach, however, is inefficient and scales poorly with template size as $wh$ distance computations are required for each query pixel. Furthermore, as nearby pixels in the template are generally close in feature space, redundant NN calculations will be made. We introduce a clustering step on the template features before the NN computation to reduce the point-set size from $wh$ to a fixed size $k$ (where $k<<wh$).  The codes in the look-up codebook are derived by performing K-means clustering and represent the $k$ major patterns present in the template. The clustering operation reduces the redundant information in template feature space, and the degree of reduction can be modulated by setting different values of $k$. The query and the template NNFs were generated by finding the nearest neighbour in the codebook for every pixel (Algorithm \ref{algo:nnf_codebook}).

While the vector quantization reduces the NN computations, it also removes the pixel-level information in the template. Hence, pixel diversity-based similarity measures and their deformable counterparts used in previous approaches \citep{talmi2017template,talker2018efficient}, which rely on all the pixels in the template and their position, can't be used. Instead, the similarity score in our method measures the shift in the quantitative and spatial distribution of the codebook patterns between the template and the query subwindow.

The template and query image NNFs are converted to one-hot encodings with $k$ categories, resulting in tensors of size $w\times h\times k$ and $W\times H\times k$, respectively. The quantitative distribution, i.e., the histogram of the codes in the template, can be computed by simply summing the tensor over all the pixels, resulting in a vector of length $k$. To calculate the histogram for all the subwindows in the query image efficiently, we convert the one-hot tensor to an integral image. The integral image was used in \citep{viola2001rapid} to efficiently gather rectangular filter responses in the object detection framework in constant time, irrespective of the window size (See Supp. Sec.~\ref{supp:int_img} for more details). The difference between the template and query histograms can be used as the similarity measure for template matching.

\subsection{Filtering in NNF}
We introduce filtering in the NNFs to encode the spatial distribution of the codebook patterns in the template, which simple histogram-based methods fail to do. Our NNF formulation represents the distribution of the patterns (collection of pixels) instead of pixels present in the images. Coarse filters operate on larger regions of an image, such as blocks or segments, rather than operating on individual pixels, making them suitable for measuring general shifts in the patterns in the images. Fig. \ref{fig:intro_fig} shows a few examples of coarse filters used in this work. Although multiple pixels in the filter kernel share the same value, the computational complexity of the filtering remains the same and scales poorly with filter size and channels. We, therefore, modify the filter kernel and implementation of the coarse filters using dilated convolutions on the integral image to work efficiently with NNFs.

\subsubsection{Modified Coarse filters}
As coarse filters operate over multiple rectangular regions within the kernel, we apply filters to the integral images for faster runtime. Dilated convolutions are often used for multi-scale context aggregation in convolutional neural networks (CNNs) for semantic segmentation\citep{yu2015multi,wei2018revisiting}. Dilated convolutions work similarly to regular convolutions but allow for skipping pixels during convolution. The number of pixels to be skipped is referred to as the dilation rate and is used to increase the receptive field of the convolution while keeping the same computational complexity.

\noindent\textbf{Dilated Convolutions:} Let $f: ([-h_{f},h_{f}] \times [-w_{f},w_{f}]) \cap \mathbb{Z}^2$ be a discrete filter of size $r_h\times r_w$, where $r_w=2w_{f}+1$, $r_h=2h_{f}+1$. The filter response is then calculated using dilated convolution as:
\begin{equation}
    (\iota \ast_{(d_x,d_y)} f)(x,y) = \frac{1}{(r_w\cdot d_x\cdot r_h\cdot d_y)} \sum_{m=-h_{f}}^{h_{f}} \sum_{k=-w_{f}}^{w_{f}} \iota\iota(x+d_x\cdot k, y+d_y\cdot m)\cdot f(k,m), 
\end{equation}

where $d_x$ and $d_y$ are the dilation rates of the kernel in the x and y direction, respectively. The receptive field of the filter is $(d_x\cdot (r_w-1)+1, d_y\cdot (r_h-1)+1)$ and hence can be modified using the dilation rate and the kernel size. For example, the receptive field of $(w,h)$ can be achieved by a convolutional kernel of size $(3,3)$ by setting the dilation rate to $((w-1)/2, (h-1)/2)$. Thus, the filter response over a larger area can be achieved while requiring only relatively few ($9$) operations for each query sub-window. This, however, comes at the cost of large bin size $(w/3, h/3)$ for the label aggregation and, thus, lower deformation granularity. This means the filtering operation can not capture distribution shifts in the NNF within the bin size. The granularity can be modulated in different ways, as discussed in Supp. Sec.~\ref{supp:filter_gm}.

\begin{figure}[tb]
\centering
\includegraphics[width=1\columnwidth]{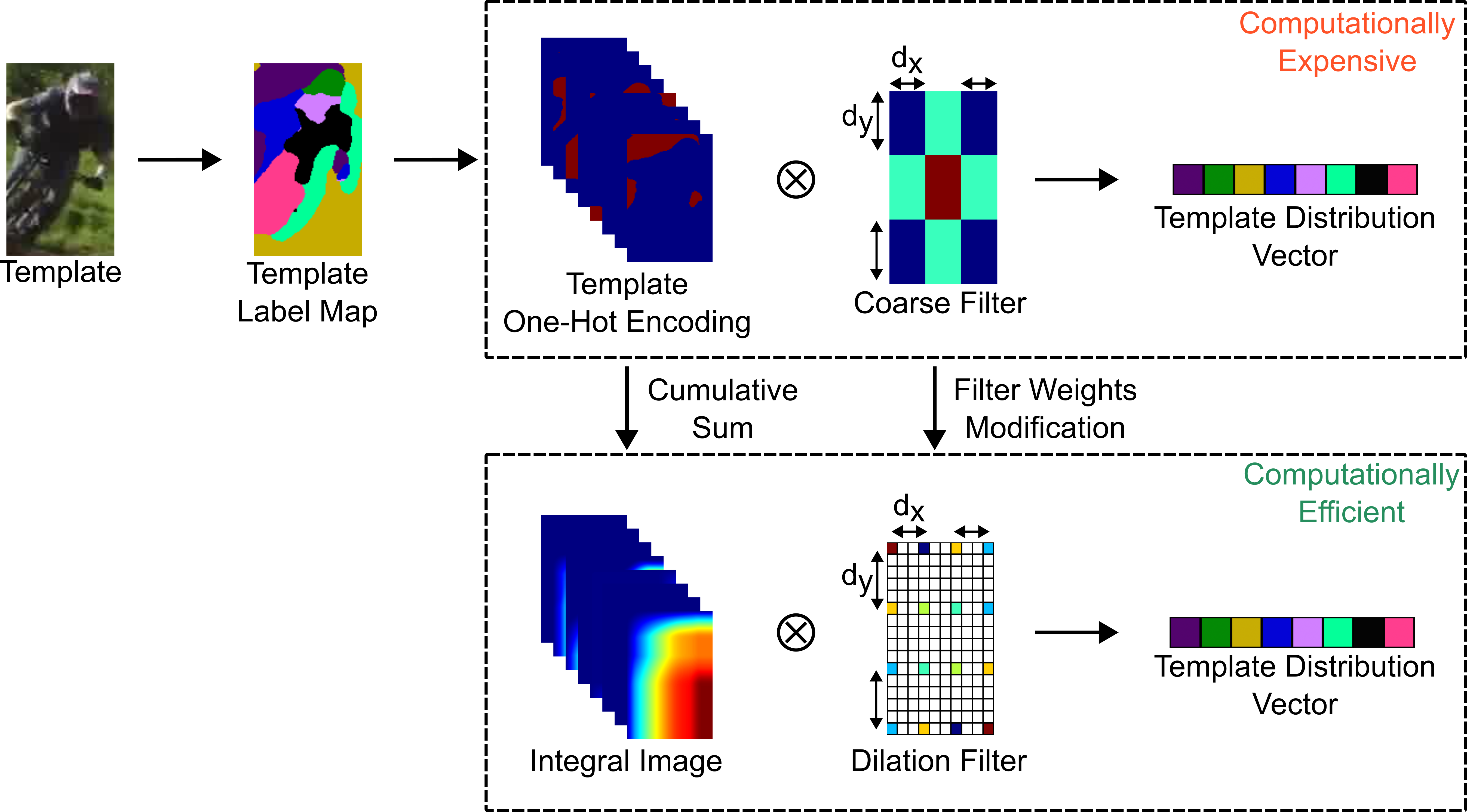}
\caption{Process to efficiently calculate the coarse filter response over the label map. The one-hot encoded image is converted to an integral image, and the coarse filter is converted into a dilation filter to reduce the computational complexity. A small $k=8$ was chosen here for illustration purposes.}
\label{fig:dil_vis}
\end{figure}

\noindent\textbf{Filter Modification for Integral Image:} The dilated filter weights were modified for coarse filter implementation. Let $f_i$ be a coarse filter of kernel size $(r_w, r_h)$ with each weight repeated over the rectangular bins of size $(d_x, d_y)$. The dilation filter $f_d$ would then have the kernel size of $(r_w+1, r_h+1)$ and the filter weights of the dilation filter can be calculated using Algorithm \ref{algo:conv_filt} (Supp. Sec.~\ref{supp:filter_wm}). Thus, the number of calculations per window has been reduced from $r_w\cdot d_x\cdot r_h\cdot d_y$ to $(r_w+1)\cdot(r_h+1)$ making the filtering process efficient. Fig. \ref{fig:dil_vis} shows the process of calculating the filter response with this approach.

\subsubsection{Encoding Spatial Information} 
As the coarse filters can now be applied to the NNFs with significantly lower computational costs, spatial information can be encoded by using multiple filters. Different filters can used to represent and compare different aspects of the template distributions, and filters can thus be chosen based on the user requirements. Here, we used the combination of coarse Gaussian and rectangular Haar-like filters to construct multiple distribution representation vectors from the filter responses. A Gaussian filter is a simple way to encode spatial information by giving more weight to the central region. Haar-like filters, introduced in \citep{viola2001rapid} for rapid object detection, encode orientation information efficiently. More complex filters can also be used to generate richer representations; however, we show that state-of-the-art performance can be achieved using these simple filters. Furthermore, the granularity of the information can be easily modified in different ways mentioned in Supp. Sec.~\ref{supp:filter_gm}. We used subscale aggregation (the c-option in Fig. \ref{fig:filter_fig}) to compare the finer shifts in the spatial distributions of the central sub-regions of the template.

\subsection{Similarity Score} 

The template representation vectors obtained from the filter responses, ${R}(\tau)$, can be defined as a collection of filtered responses:
\begin{gather*}
    {R}(\tau)= \{ R_{s,f}(\tau) \} \\
    \text{where, } R_{s,f}(\tau) =  \iota\iota_{\tau} \ast_{s} f , \forall s \in S, f \in \mathcal{F} 
\end{gather*}
\noindent where, $S=\{1,2,...,s\}$ defines the set of all levels of sub-scale aggregation considered for the template, i.e., for a scale $s$, the dilation rate was changed such that the receptive field of the filter was $(w/s, h/s)$. $\iota\iota_{\tau}$ represents the integral image of the template NNF, and $\mathcal{F}= \{f_i\}_{i=1}^{n}$ represents the set of $n$ different filters used for the distribution representation. The number of template representation vectors would then be $n\cdot S$. The distribution similarity score at each sub-window $q_i \in Q$ in the query image is then defined as
\begin{equation}
     \text{sim}(q_i, \tau) =  \sum_{s\in S} \sum_{f \in \mathcal{F}} -w_{s,f} \cdot |R_{s,f}(\tau) - R_{s,f}(q_i)|
\end{equation}
where $w_{s, f}$ is the weight assigned to the difference in the filter response of the filter $f$ at scale $s$.

\section{Computational Complexity}

\noindent\textbf{K-means computation:} The first step in the method is to calculate the $k$ cluster centers of the $w \times h \triangleq l$ template points. The k-means clustering algorithm complexity is $O(klt)$, where $t$ is the number of iterations. The complexity would be $O(l)$ as ${k, t}\ll l$.

\noindent\textbf{NN Search:} The NN calculation is done by calculating the euclidean distance of all the points in the image $W\times H\triangleq L$ to the $k$ cluster centers. The complexity of this step is $O(kL)$.

\noindent\textbf{Filter Responses:} The indices of the NN are converted into a one-hot tensor of size $W \times H \times k$ and an integral image is constructed in $O(L)$. Different coarse filters at different scales can be applied to aggregate the distribution. Each filtering operation with the dilated filter size $(r_h+1)\times (r_w+1)$ has a time complexity of $O(((r_h+1)\cdot (r_w+1)\cdot k\cdot L)$. Multiple scale representations are then computed in $O(s\cdot (r_h+1)\cdot (r_w+1)\cdot k\cdot L)$ where $s$ is the maximum number of scales considered. Similarly, adding $n$ different filters would increase the complexity by $O(n)$. Here $r_h\cdot r_w$ can be ignored as $r_h\cdot r_w \ll k,L$. Hence, the overall complexity of this step would be $O(nskL)$.

\noindent\textbf{Similarity Score Calculation:} The distribution similarity score is calculated by subtracting the $n\cdot s$ template distribution vectors from the query distribution vector in $O(nskL)$. 

\noindent\textbf{Target Localization:} Maxima location requires another sweep through the image which takes another $O(L)$ computations.

Overall, the complexity of our method would be $O(l)+O(kL)+O(nskL)+O(nskL)+O(L)$ $\approx O(nskL)$ where $ 1 < s\cdot n\cdot k < l \ll L$. Furthermore, the operations are easily parallelizable in a GPU to speed up the computations.

\section{Experiments}\label{sec:exps}
\subsection{Datasets}
We evaluated our method on three datasets with different image sizes and challenges. The first dataset, BBS, a subset of the Online Object Tracking benchmark \citep{wu2013online}, consists of 90 video sequences with challenges such as complex deformation, occlusions, scale differences, etc. It was acquired by sampling frames from \citep{wu2013online} at different intervals. Three random pairs of frames with constant frame differences, $dF=\{25, 50, 100\}$, were extracted from each sequence. This resulted in three sub-datasets, namely, BBS25, BBS50, and BBS100, consisting of 270, 270, and 252 images, respectively. The sampling was repeated five times to extract more robust statistics and simulate template matching in the wild more closely. The image size in the dataset was relatively small, 320x480 or 320x240 pixels.

The TinyTLP dataset is based on a shortened version of the object-tracking dataset Track Long and Prosper (TLP) \citep{moudgil2018long}. This dataset contains 50 video clips with 600 frames of size 1280x720 pixels. To reduce redundancy, we followed the protocol used in \citep{talker2018efficient} and sampled 50 frames [1, 11, ..., 491] from each video and then used the 100th next frame as the query pair. The dataset created consists of 2500 template-query pairs. 

The TLPattr dataset \citep{moudgil2018long} is a collection of 91 short clips of different durations focusing on six different challenge attributes in the TLP dataset, namely, illumination variation, fast motion, background clutter, out-of-view or full occlusions, scale variations, and partial occlusions. The sequences in TLPattr are selected such that only one of the abovementioned challenge attributes is present in a sequence. There are 15 sequences belonging to a particular attribute, except for scale variation, which has 16 sequences. We randomly selected 15 images from each sequence as templates and chose the 100th next frame as the query image. Overall, the dataset consists of 1351 template-query pairs.

\subsection{Implementation}
We implemented our algorithm in \verb|python| programming language using \verb|PyTorch| \citep{NEURIPS2019_9015} library to utilize GPU parallelization efficiently. All experiments were performed on an \verb|AMD Ryzen 9 3900X| CPU and a \verb|NVIDIA GeForce RTX 2080 Ti| GPU, respectively. The code for our method and the dataset generation is available on GitHub: \url{https://github.com/aktgpt/vq-nnf}.

\noindent\textbf{Feature Extraction:} Following a similar feature extraction strategy as in \citep{talmi2017template,talker2018efficient}, we investigated and compared two feature descriptors in our experiments: colour and deep features. The colour features were extracted by taking the 3x3 overlapping patches of RGB values for each pixel, resulting in a total of 27 features per pixel. The deep features were extracted from a pre-trained ResNet-18 \citep{he2016deep}. This differs from \citep{talmi2017template,talker2018efficient} where the VGG model is used. ResNet-18 is a smaller and more parameter-efficient alternative with slightly lower performance on the ImageNet dataset (69.7\%, 89.1\% vs 72.4\%, 90.8\% top-1 and top-5 accuracy) but with a significantly lower number of parameters (11.7 million vs 143.7 million). Here, the features were extracted from the bottleneck and consecutive stages in the model, concatenated and resized to the original input. The highest feature dimension considered in our experiments was 512, for which the output from the \verb|conv1| (64), \verb|conv2_x| (64), \verb|conv3_x| (128), and \verb|conv4_x|  (256) layers were concatenated and resized to the original image size.

\noindent\textbf{Codebook and NNF Generation:}
We used a fast GPU PyTorch-based clustering library\footnote{The k-means clustering library is available at:\url{https://github.com/DeMoriarty/fast_pytorch_kmeans}} to perform simple K-means clustering to find $k$ cluster centers to represent the template codebook. We then found the exact NN for each feature vector in the template and the query image in the $k$ template cluster centers based on the L2 distance measure. The NNFs were then converted to one-hot encoding, and integral images were calculated for filtering. We explored different values of $k$ in our experiments (Fig. \ref{fig:tlp_ablation}, and Supp. Fig. \ref{fig:bbs_ablation} ) and chose $k=128$ for our main experiments.

\noindent\textbf{Filtering:} In our experiments, we used a $3\times 3$ Gaussian kernel with a sigma of 2 and dilation of $(w/3, h/3)$ to give equal weights to all the bins. This was intentional; we wanted to model the histogram comparison without any spatial location preferences and treat the performance as the baseline. For modelling the spatial distribution shifts, we experimented with different sets of filters, namely $\mathcal{F}_{gauss}$ using only Gaussian filter, $\mathcal{F}_{2-rect}$ containing haar-2x and haar-2y filters, $\mathcal{F}_{3-rect}$ containing haar-3x, and haar-3y filters and $\mathcal{F}_{2,3-rect}$ containing both 2-rectangle and 3-rectangle filters. The filter kernel was multiplied with the Gaussian kernel. The Haar filter sets also contain the Gaussian filter. The weight for different filters $f$ at scale $s$ ($s \in S$), $w_{s,f}$ was empirically set to $(1/s)\cdot w_f$, where $w_f$ is set to 1 for Gaussian filter and 0.25 for Haar filters. We found that in the BBS dataset, the best overall performance was achieved at $k=128$, $S=3$, with filter sets $\mathcal{F}_{2,3-rect}$ (overall 15 filters) and $\mathcal{F}_{2-rect}$ (9 filters) for color and deep features respectively. Our best configuration in the TLP datasets was with 
$k=128, \mathcal{F}_{2,3-rect}$ with the scale of $S=4$ (20 filters) and $S=3$ (15 filters) for color and deep features, respectively.

\noindent\textbf{Quantitative Evaluation:} We evaluated the performance of the different representation schemes in our method using Intersection over Union (IoU) between the estimated window, $\tau_x$ from the algorithm and the ground truth, $\tau_{GT}$. It is defined as follows:

\begin{equation}
    IoU(\tau_{x},\tau_{GT}) = \frac{\tau_{x} \cap \tau_{GT}}{\tau_{x} \cup \tau_{GT}}.
\end{equation}

The performance for each dataset is shown as the mean IoU (MIoU), i.e. the mean of IoUs of all the samples, and Success Rate (SR), which is defined as the fraction of the samples in the dataset where $IoU > 0.5$. 

\noindent\textbf{Performance Comparison:} We compare our method with the official code releases of DDIS \citep{talmi2017template} and DIWU \citep{talker2018efficient} as they are closest approaches to ours. We also compare with DIM \citep{spratling2020explaining} and Deep-DIM \citep{gao2022shape} in our main result to show the comparison with a non-NNF-based benchmark on colour and deep features, respectively. The reported time in this work refers to the time the algorithm takes to generate the query similarity heatmap after feature extraction. For NNF-based methods, this differs from the reported time in the \citep{talmi2017template,talker2018efficient} where the NNF creation time is excluded. The computation time of our method is not directly comparable as the computations for our method are performed on a GPU, and the official implementation of these methods uses a CPU for NNF computation. However, the relative runtime difference between the methods using color vs. deep features highlights the challenges in using DDIS and DIWU effectively, especially at higher resolutions. Note that for Deep-DIM \citep{gao2022shape}, the original model trained by the authors with stylized data was used. 

\emph{Note:} Even though all the Python random seeds were fixed in the experiments, the inherent system randomness might cause the evaluation metric results to differ in different experimental runs. For fair evaluation, each ablation experiment is conducted within the same script run to minimize such randomness within the experiment. This also showcases the spread of the performance of our method and indicates the possible randomness when using it in real settings.

\subsection{Results}

The results for the BBS25, BBS50, and BBS100 datasets are shown in Tab. \ref{tab:bbs_final}. The images in the datasets are relatively small (320x480 or 320x240). Our method lags slightly behind DDIS and DIWU for color features; however, for the deep features, the performance is marginally better than that of others. Our method outperforms DIM and Deep-DIM for both color and deep features. The runtime of our best method for the deep features is approximately the same as the color features, while the runtime increased around 4.7 times for DDIS (0.349 to 6.337 sec) and 16.8 times for DIWU (0.349 to 5.883). This further shows that increasing the feature dimensions in our method provides better performance without compromising the speed.

\begin{table*}[tb]
\caption{Performance results for BBS datasets. color features are marked with (C), and ResNet features are denoted by (D). The methods run on GPU are denoted by (G). The best results in each column are written in bold.}
\centering
\resizebox{1\columnwidth}{!}{
\vspace{4mm}
\begin{tabular}{c|c|c|c|c|c|c|c|c|c|}
\hline
\multicolumn{1}{|c|}{\textbf{Method}} & \multicolumn{2}{|c|}{\textbf{BBS25}} & \multicolumn{2}{|c|}{\textbf{BBS50}} & \multicolumn{2}{|c|}{\textbf{BBS100}} & \multicolumn{3}{|c|}{\textbf{Total}} \\ \hline
\multicolumn{1}{|c|}{} & SR & MIoU & SR & MIoU & SR & MIoU & SR & MIoU & Time \\ \hline
\multicolumn{1}{|c|}{DDIS (C)} & \textbf{0.781} & \textbf{0.638} & \textbf{0.695} & \textbf{0.567} & \textbf{0.594} & \textbf{0.493}& \textbf{0.690} & \textbf{0.566} & 1.336\\ \hline
\multicolumn{1}{|c|}{DIWU (C)}  & 0.771 & 0.624 & 0.684 & 0.554 & 0.584 & 0.485 & 0.680 & 0.554 & 0.349\\ \hline
\multicolumn{1}{|c|}{DIM (C)}  & 0.743 & 0.621 & 0.633 & 0.525 & 0.542 & 0.458 & 0.639 & 0.535 & 6.935\\ \hline
\multicolumn{1}{|c|}{Ours best (C) (G)} &  0.759 & 0.632 &  0.651 & 0.545 &  0.556 &  0.476 &  0.655 &  0.551 & \textbf{ 0.114}\\ \hline \hline

\multicolumn{1}{|c|}{DDIS (D)} & 0.813 & 0.667 & 0.748 & \textbf{0.613} & 0.682 & 0.569 & 0.748 & 0.617 & 6.337\\ \hline
\multicolumn{1}{|c|}{DIWU (D)}  & \textbf{0.821} & 0.665 & \textbf{0.751} & 0.611 & \textbf{0.690} & 0.571 & \textbf{0.754} & 0.616 & 5.883\\ \hline
\multicolumn{1}{|c|}{Deep-DIM (D) (G) }  & 0.800 & 0.652 & 0.726 & 0.593 & 0.640 & 0.533 & 0.722 & 0.593 & 25.404 \\ \hline
\multicolumn{1}{|c|}{Ours best (D) (G)} & 0.819 & \textbf{0.679} & 0.749 &  0.614 & \textbf{0.694} & \textbf{0.579} & \textbf{0.754} & \textbf{0.624} & \textbf{0.119}\\ \hline \hline

\end{tabular}%
}
\label{tab:bbs_final}
\end{table*}

The results for the TinyTLP and TLPattr datasets are presented in Tab.~\ref{tab:tlp_final}. The images in the datasets are of higher resolution (1280x720), which makes the datasets suitable for showcasing the advantages provided by our method compared with DDIS and DIWU. DIM and Deep-DIM were not considered for the TinyTLP dataset due to runtime. For color features, our best configuration outperforms the DDIS by approx. 0.9\% and 1.2\% in TinyTLP and TLPattr datasets, respectively. Our method outperforms DIWU by 1.5\% on the TinyTLP dataset but lags by 1\% in the TLPattr dataset. DIM was the worst-performing method, with 3.1\% lower MIoU compared to ours. For deep features, our best configuration outperforms the DDIS by approx. 4.7\% and 2.7\%, and DIWU by approx. 6.1\% and 2.5\% in TinyTLP and TLPattr datasets, respectively. Our method outperforms Deep-DIM by 5.9\% on the TLPattr dataset. Here, similar to the BBS performance, the color and deep features runtimes of our method are marginally different for both datasets. In contrast, for DDIS, the runtime increased by 2.2 (22.5 to 49.7 secs) and 1.6 (25.5 to 41.4 secs) times for TinyTLP and TLPattr, respectively, and for DIWU, the runtime increased by 13.6 (2.5 to 33.9 secs) and 17.5 times (1.9 to 34.1 secs) for TinyTLP and TLPattr, respectively.

\begin{table*}[tb]
\caption{Performance results for TinyTLP and TLPattr datasets. color features are marked with (C), and ResNet features are denoted by (D). The methods run on GPU are denoted by (G). The best results in each column are written in bold.}
\centering
\resizebox{0.8\columnwidth}{!}{
\vspace{4mm}
\begin{tabular}{c|c|c|c|c|c|c|}
\hline
\multicolumn{1}{|c|}{\textbf{Method}} & \multicolumn{3}{|c|}{\textbf{TinyTLP}} & \multicolumn{3}{|c|}{\textbf{TLPattr}}  \\ \hline
\multicolumn{1}{|c|}{} & SR & MIoU & Time & SR & MIoU & Time \\ \hline
\multicolumn{1}{|c|}{DDIS (C)} & 0.607 & 0.528 & 22.565 & 0.557 & 0.479 & 25.486 \\ \hline
\multicolumn{1}{|c|}{DIWU (C)} & 0.613 & 0.522 & 2.508 & \textbf{0.600} & \textbf{0.501} & 1.948 \\ \hline
\multicolumn{1}{|c|}{DIM (C)} & - & - & - & 0.537 & 0.460 & 114.839 \\ \hline
\multicolumn{1}{|c|}{Ours best (C) (G)} & \textbf{0.614} & \textbf{0.537} & \textbf{0.511} & 0.576 & 0.491 & \textbf{0.563} \\ \hline \hline

\multicolumn{1}{|c|}{DDIS (D)} & 0.631 & 0.549 & 49.686 & 0.617 & 0.522 & 41.362\\ \hline
\multicolumn{1}{|c|}{DIWU (D)} & 0.619 & 0.535 & 33.899 & 0.628 & 0.524 & 34.063\\ \hline
\multicolumn{1}{|c|}{Deep-DIM (D) (G)} & - & - & - & 0.574 & 0.490 & 25.716\\ \hline
\multicolumn{1}{|c|}{Ours best (D) (G)} & \textbf{0.692} & \textbf{0.596} & \textbf{0.582} & \textbf{0.652} & \textbf{0.549} & \textbf{0.579} \\ \hline \hline

\end{tabular}%
}
\label{tab:tlp_final}
\end{table*}


Following the quantitative results in the previous section, we also show the qualitative comparison of our method with DDIS and DIWU. Fig. \ref{fig:tlp_vis_comp} and \ref{fig:bbs_vis_comp} (In Supplementary Sec.~\ref{supp:bbs_vis}, Fig.~\ref{fig:bbs_vis_comp}) show the results of the methods on the BBS and TLPattr datasets, respectively. Although the heatmaps appear noisier than the compared methods, our method successfully finds the template in the query image. Furthermore, our method seems robust in cases where objects similar to the template are present while other methods fail to detect the template successfully.  

Since DIM and Deep-DIM show lower performance on all the datasets, we haven't considered them in further experiments.

\begin{figure}[tb]
\centering
\includegraphics[width=1\columnwidth]{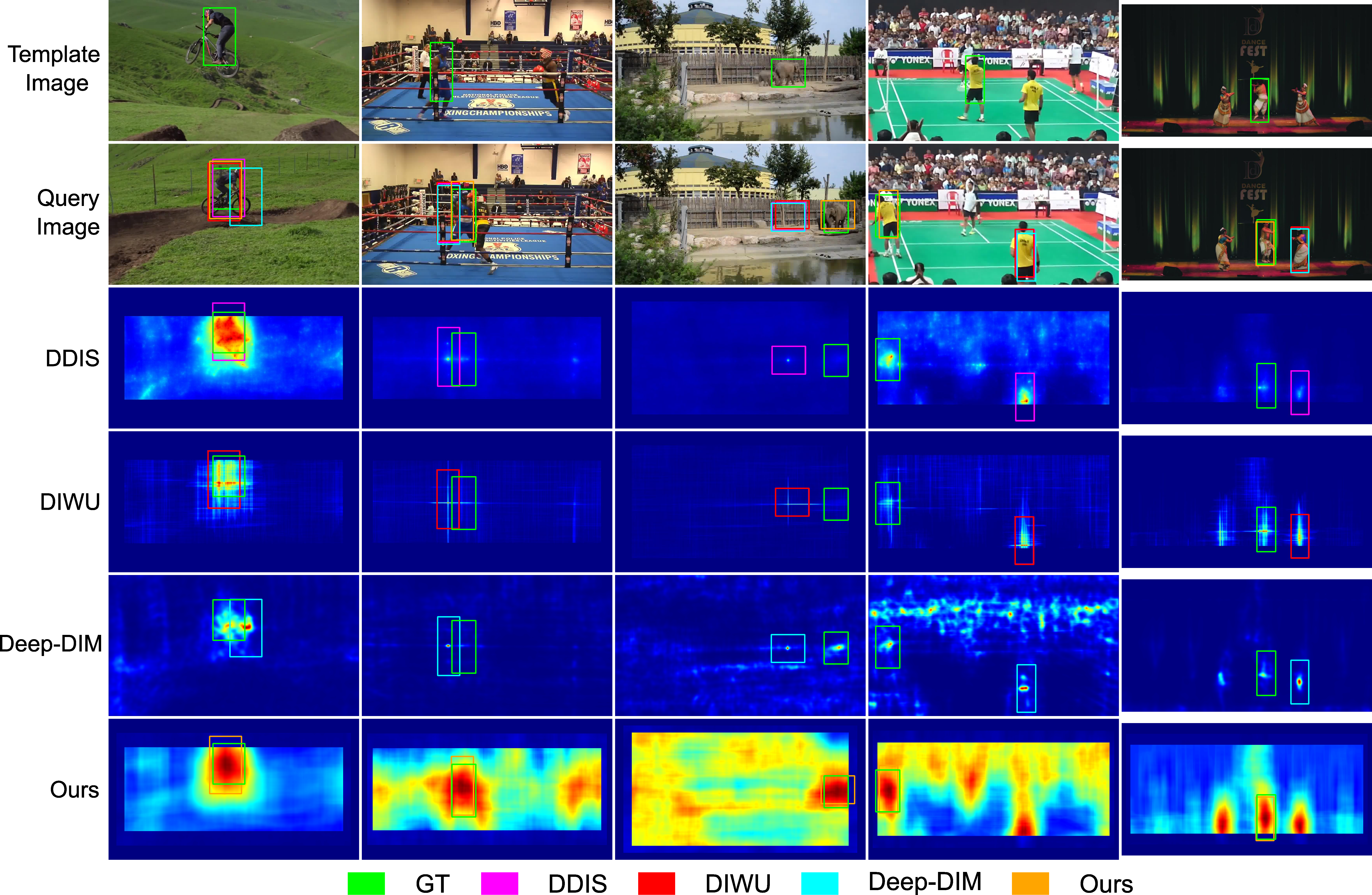}
\caption{Qualitative results for the TLPattr dataset using deep features. The first column shows the template image with the template marked in green. The second column shows the query image with the results of DDIS (in pink), DIWU (in red), Deep-DIM (in blue) and our method (in orange). The third to sixth columns show the heatmaps of the DDIS, DIWU, Deep-DIM, and our method, respectively, with the predicted bounding box marked in the same colours as the second column.}
\label{fig:tlp_vis_comp}
\end{figure}

\subsection{Scale Experiments}

We compare the effect on performance and runtime with respect to the image resolution on the TLPattr dataset. The original resolution is $1280\times 720$; hence, we downsample the images with different scale factors in the $[1/4, 1]$ range. The template, query images, and ground truths are downsampled with the scale factor. Scale factors above one are not chosen as the image resolution is high enough to showcase the runtime variations, and upsampling doesn't add any meaningful information. The experiment also reflects the robustness of the performance if downsampling of the images is preferred to speed up the process. 

\begin{figure}[tb]
\centering
\includegraphics[width=1\columnwidth]{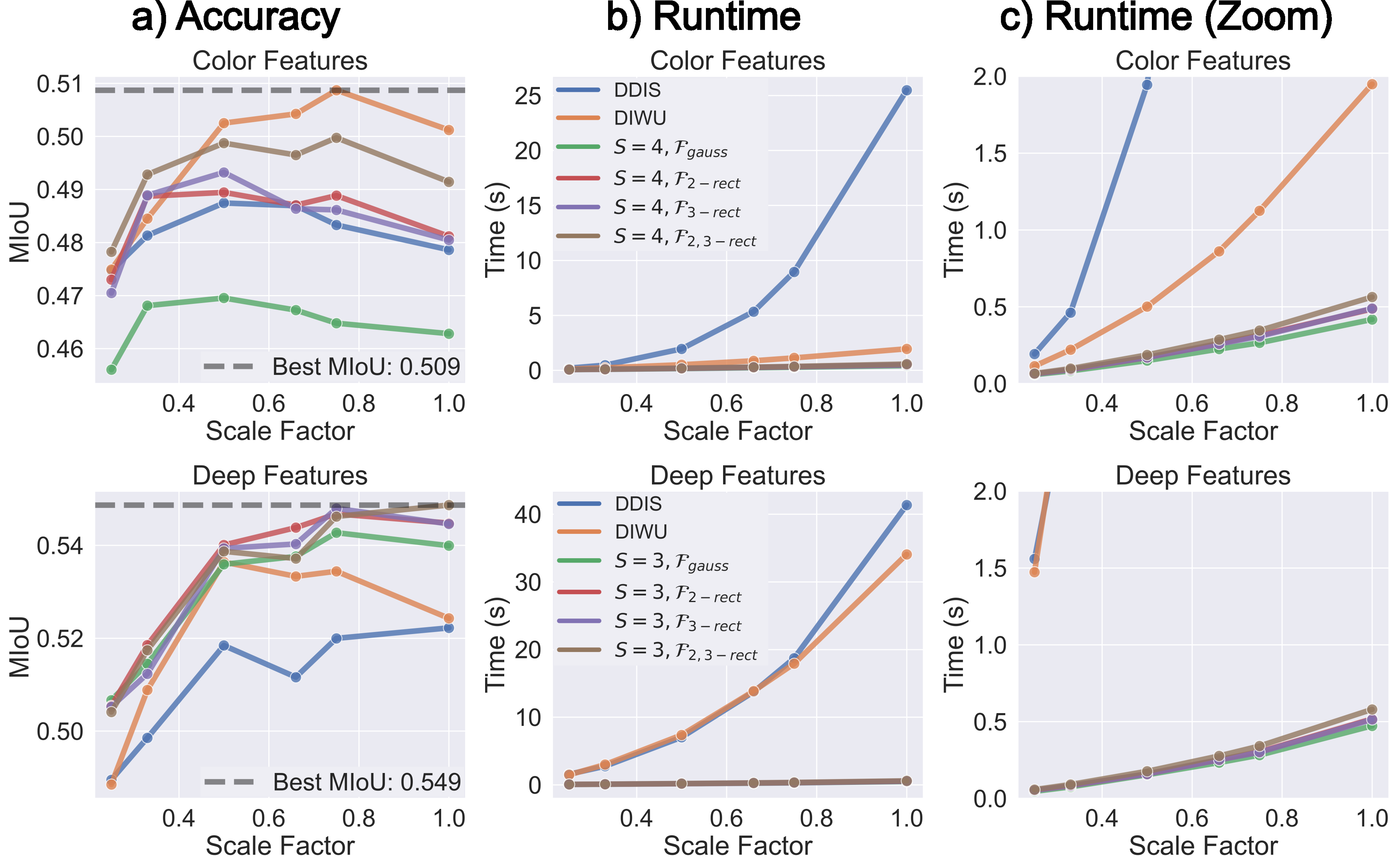}
\caption{MIoU and Runtime performance at different scale factors on the TLPattr dataset. The x-axis shows the relative scale factor with the image of size $1280\times 720$. The y-axis corresponds to a) The mean IoU of the algorithms with the color features (top) and deep features (below), b) the runtime of the algorithms and c) the zoomed-in runtime with the y-axis clipped to two seconds. }
\label{fig:scale_comp}
\end{figure}

Figure \ref{fig:scale_comp} shows our method's MIoU and runtime performance along with DDIS and DIWU. As can be seen from the color feature accuracy plot, our methods' performance peaked at a lower resolution (between 0.5-0.75) and either remained the same or decreased. We hypothesize that color features provide limited benefits at higher resolutions as all the methods show a downward trend towards larger scale factors. Our methods' performance for color features remains between DDIS and DIWU, except for $\mathcal{F}_{gauss}$, which was the worst-performing configuration. For deep features, all configurations of our method perform better than DDIS and DIWU. Furthermore, after the scale factor of 0.5, our methods show only fractional change in performance, which indicates that images can be processed at a lower resolution while maintaining performance. 

The runtime analysis is shown in Fig.~\ref{fig:scale_comp} b), and c). As can be seen from the figures, all configurations of our method scale better with the image resolution than DDIS and DIWU. For color features, the DIWU scales similarly to our methods with the resolution, but with deep features, both DDIS and DIWU scale equally poor with resolution.

\subsection{Attribute Experiments}

\begin{figure}[tb]
\centering
\includegraphics[width=1\columnwidth]{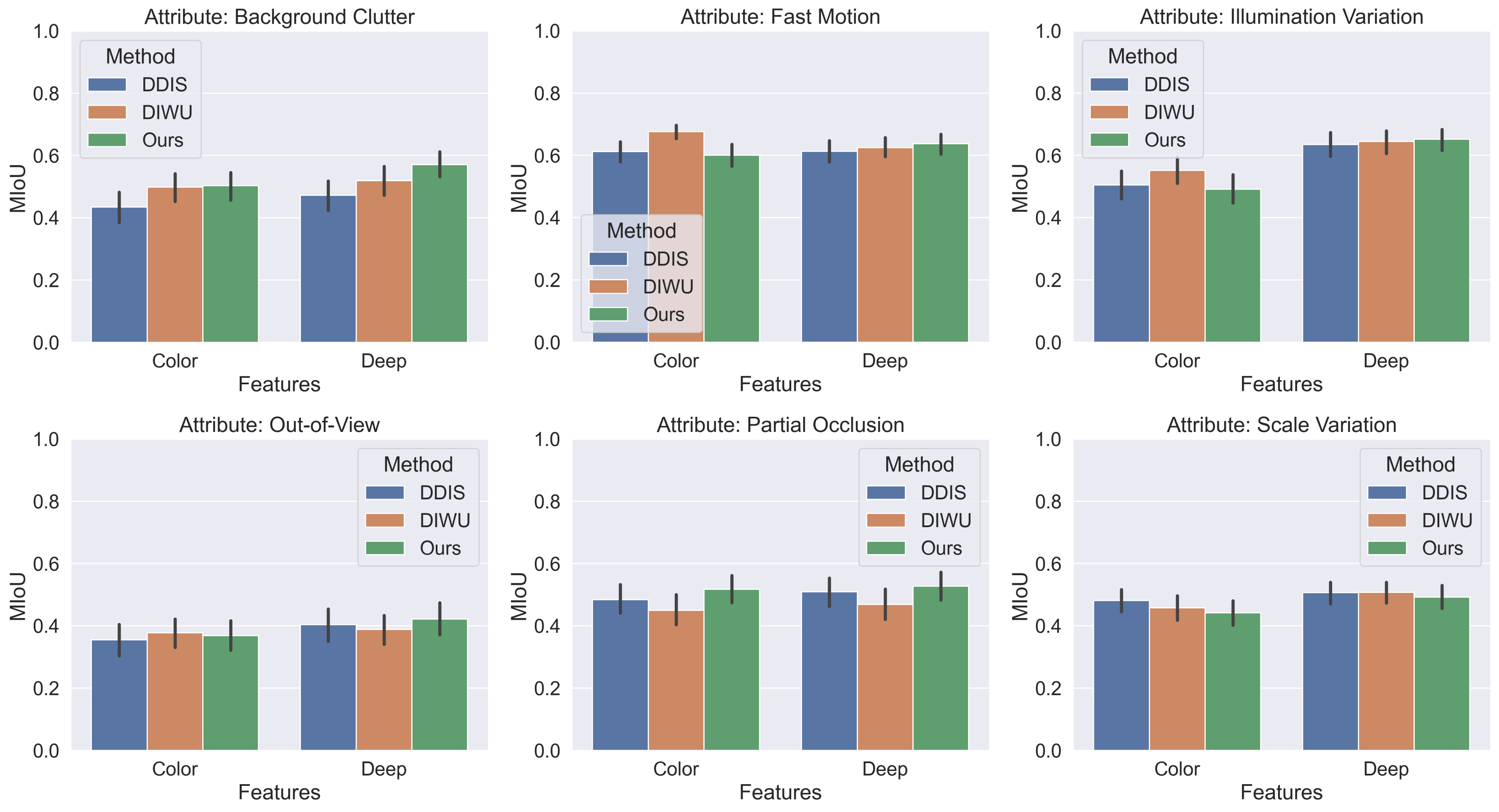}
\caption{The MIoU performances of our best method with the DDIS and DIWU for the different challenge attributes present in the TLPattr dataset. DDIS, DIWU and our performance are shown in blue, yellow and green bars, respectively.}
\label{fig:attr_comp}
\end{figure}

We evaluate the performance of our method on the various challenge attributes of the TLPattr dataset and again compare it with DDIS and DIWU. The attribute-wise performance for both color and deep features is shown in Fig. \ref{fig:attr_comp}. Our best method performs comparably to DDIS and DIWU for most of the challenge attributes. DDIS and DIWU rely on a smaller subset of good matches, making them robust to partial occlusions and background clutter. In contrast, our method relies on the distribution of all the matches. Our method outperforms them in both categories, which shows that distribution matching might be better at a higher resolution than NN diversity-based scoring. For the rest of the categories, our performance is similar to or slightly better than DDIS and DIWU.

\subsection{Dimensionality Reduction}

Although our method makes the application of high-dimensional features relatively fast over the GPU, using a lower-dimensional representation for the NNF creation might still be desirable for lower memory and shorter runtime requirements. DDIS and DIWU use PCA dimensionality reduction to reduce the computational time of the methods. The methods reduce the dimensionality of the features to $D=9$ first to speed up the NN search. Here, similarly, we explore the effects of dimensionality reduction by reducing the feature dimensions of our deep model from $D=512$ to $D=9$ and $18$ and measure the relative change in performance on all the datasets. Only the results of our best-performing configuration on the datasets are shown.

\begin{table*}[tb]
\caption{Dimensionality reduction results for the BBS and TLP datasets. The best results in each column are written in bold.}
\centering
\resizebox{1\columnwidth}{!}{
\vspace{4mm}
\begin{tabular}{c|c|c|c|c|c|c|}
\hline
\multicolumn{1}{|c|}{\textbf{Feature Dim.}} & \multicolumn{4}{|c|}{\textbf{BBS Datasets}} & \multicolumn{2}{|c|}{\textbf{TLP Datasets}}  \\ \hline
\multicolumn{1}{|c|}{} & \textbf{BBS25} & \textbf{BBS50} & \textbf{BBS100} & \textbf{BBS(Total)} & \textbf{TinyTLP} & \textbf{TLPattr}  \\ \hline
\multicolumn{1}{|c|}{MIoU(D=512)} & 0.674 & \textbf{0.615} & \textbf{0.579} & \textbf{0.622} & \textbf{0.596} & \textbf{0.549} \\ \hline
\multicolumn{1}{|c|}{MIoU(D=18)} & \textbf{0.679} & 0.610 & 0.578 & \textbf{0.622} & 0.592 & 0.547 \\ \hline
\multicolumn{1}{|c|}{MIoU(D=9)} & 0.671 & 0.607 & 0.566 & 0.614 & 0.587 & 0.548 \\ \hline \hline

\multicolumn{1}{|c|}{Time(D=512)} & 0.123 & 0.124 & 0.121 & 0.122 & 0.582 & 0.579\\ \hline
\multicolumn{1}{|c|}{Time(D=18)}  & \textbf{0.096} & \textbf{0.096} & \textbf{0.097} & \textbf{0.096} & 0.432 & 0.480 \\ \hline
\multicolumn{1}{|c|}{Time(D=9)} & \textbf{0.096} & 0.097 & 0.098 & 0.097 & \textbf{0.424} & \textbf{0.478} \\ \hline \hline

\end{tabular}%
}
\label{tab:pca}
\end{table*}

Tab. \ref{tab:pca} shows the MIoU and runtime of the dimensionality reduction on all the datasets. The overall performance of our method on the BBS datasets is reduced by approx. 1\% of the original when using the same dimensions ($D=9$) as used in the DDIS and DIWU. The runtime of our method, however, was reduced by approx. 20\%. For $D=18$, a similar performance to the original was achieved while reducing the runtime by a similar margin.

For TLP datasets, a similar performance reduction of approx. 1\% was noticed for $D=9,18$. This shows the robustness of the distribution matching on high-resolution datasets, even with dimensionality reduction. The runtime for $D=9$ was reduced by approx. 27\% and 17\% for TinyTLP and TLPattr datasets, respectively.

\subsection{Design Choice Evaluation}
\begin{figure}[tb]
\centering
\includegraphics[width=1\columnwidth]{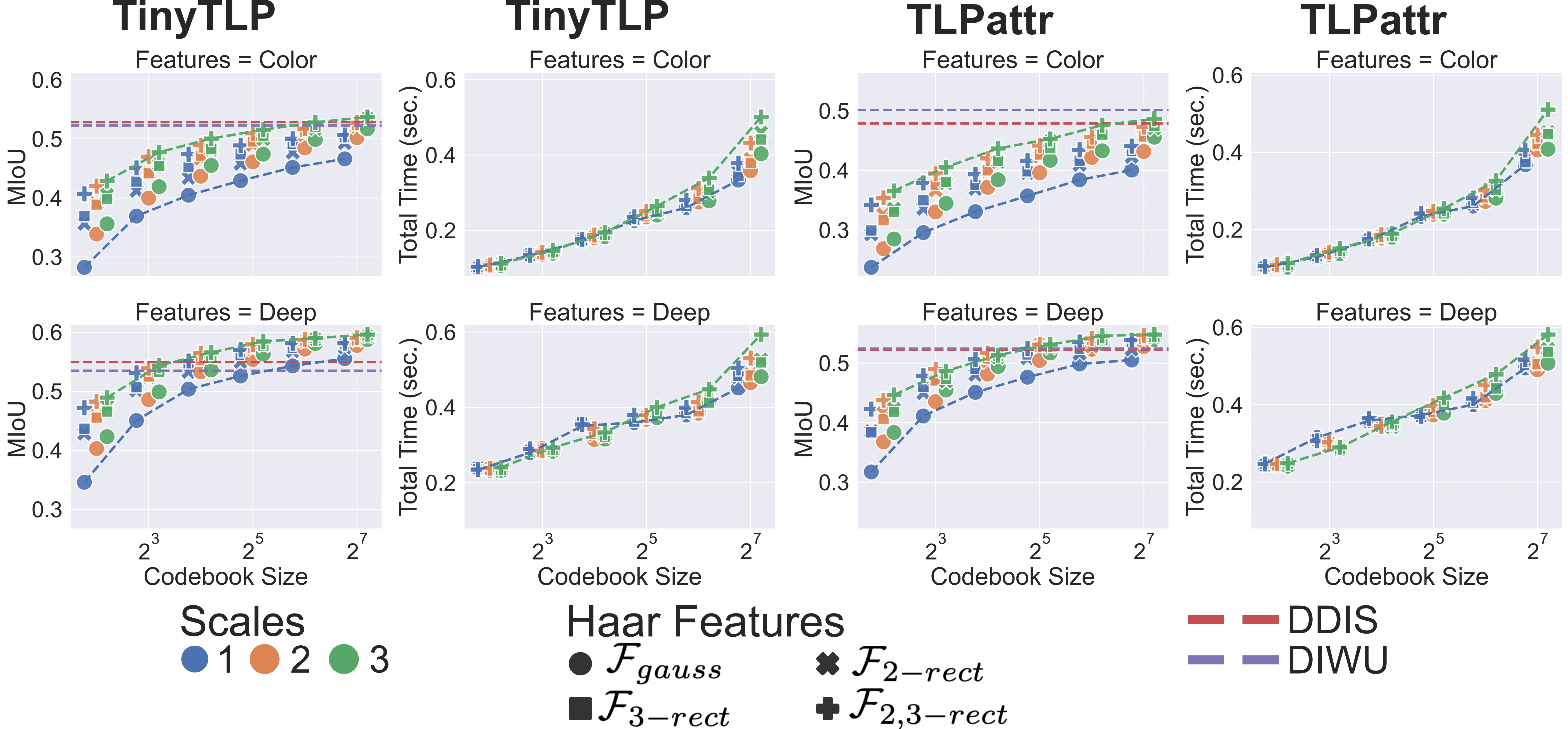}
\caption{Average MIoU (top two rows) and Total time (bottom two rows) performance of our approach with different hyper-parameters on the TinyTLP and TLPattr datasets. The different scales shown in different colours are shifted slightly for better visualization. DDIS and DIWU performances are shown with the dashed lines in the MIoU plots. The lines in the plots highlight the trends for the base configuration without spatial information (blue line) and the configuration with the most filters and scales (green line) for each cluster size.  The x-axis for all the plots is shown in the log scale.}
\label{fig:tlp_ablation}
\end{figure}

\begin{figure}[!tb]
\centering
\includegraphics[width=0.85\columnwidth]{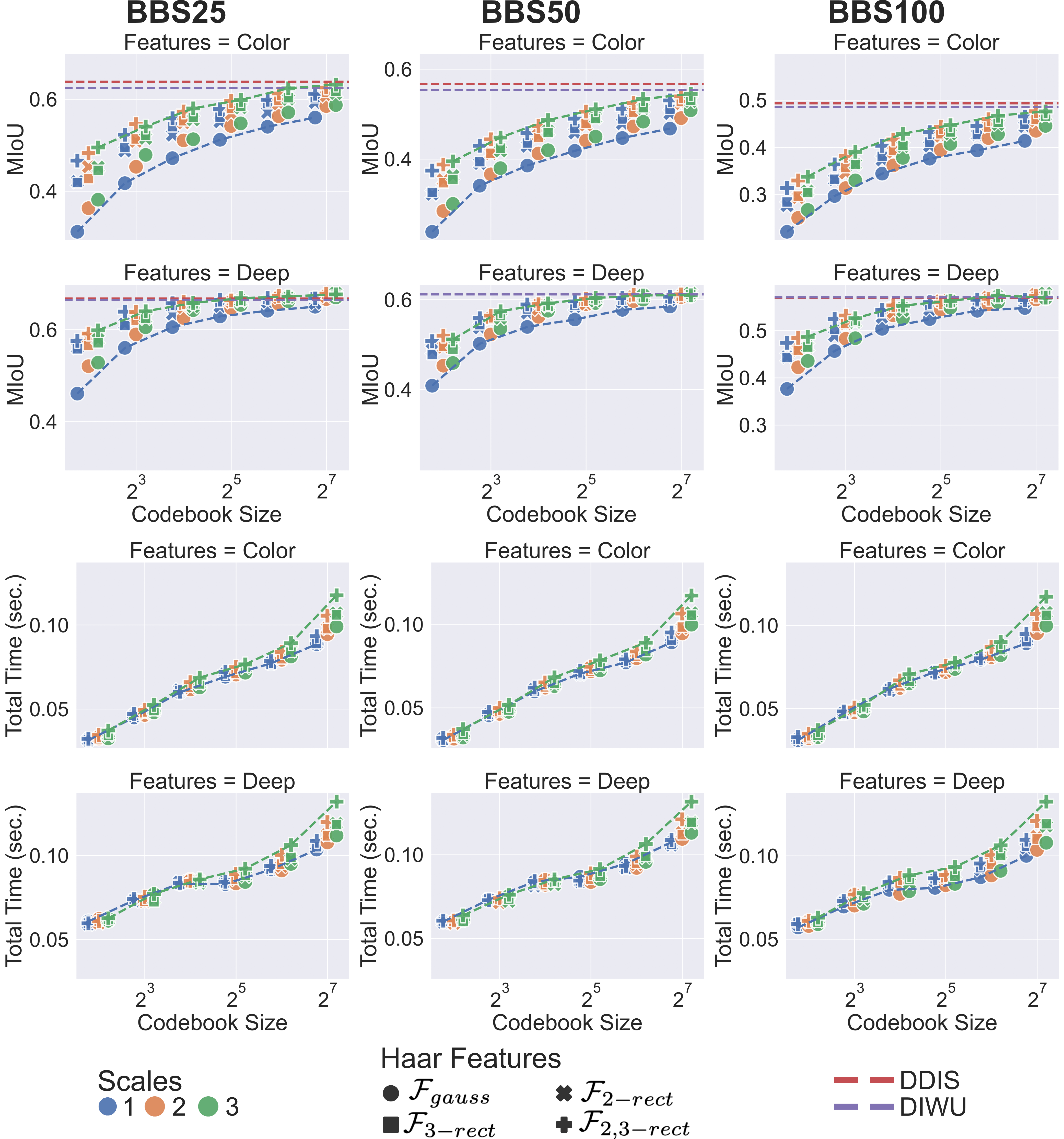}
\caption{Average MIoU (top two rows) and Total time (bottom two rows) performance of our approach with different hyper-parameters on the BBS25, BBS50, and BBS100 datasets. The different scales shown in different colours are shifted slightly for better visualization. DDIS and DIWU performances are shown with the dashed lines in the MIoU plots. The lines in the plots highlight the trends for the base configuration without spatial information (blue line) and the configuration with the most filters and scales (green line) for each cluster size.  The x-axis for all the plots is shown in the log scale.}
\label{fig:bbs_ablation}
\end{figure}

We evaluate the performance of our method on the datasets for different hyperparameters, codebook sizes (k), multiple distribution aggregation scales (S), and different filter sets. We considered color and deep features with feature dimensions, $d=\{27,512\}$, number of clusters, $k= \{4,8,16,32,64,128\}$, total scales, $S=\{1,2,3\}$, and the filter sets ($\mathcal{F}_{gauss}$, $\mathcal{F}_{2-rect}$,$\mathcal{F}_{3-rect}$,$\mathcal{F}_{2,3-rect}$). Fig.~\ref{fig:tlp_ablation} shows the MIoU and run-time performance on TinyTLP and TLPattr datasets. Fig.~\ref{fig:tlp_ablation} shows the method's performance on the TLP datasets to highlight the advantage of our method in high-resolution data. The performance of the choices on the BBS datasets (low-resolution) is shown in Fig.~\ref{fig:bbs_ablation} (Supp. Section \ref{supp:bbs_dce}). Below, we discuss the trends seen from the figures and the impact of different parameters used in our algorithm.  

\textit{Codebook Size.} In our method, the codebook contains $k$ patterns present in the template. Increasing the codebook size will increase the representation capacity of the codebook and, thus, performance, as seen in each plot (increasing codebook size on the x-axis of MIoU plots) and between plots (increasing feature dimensions from 27 to 512) in Fig.~\ref{fig:tlp_ablation},~\ref{fig:bbs_ablation}. Furthermore, even without including any spatial information (i.e., the base method, $K=128, S=1, \mathcal{F}_{gauss}$), our method shows the competitiveness of simple histogram matching in the NNFs with the similarity measures used in DDIS and DIWU. VQ formulation allows for a flexible way of choosing the $k$ based on the user requirements when using deep features. Intuitively, $k$ can be chosen based on the template size; as deep features encode a larger region in the representation, a smaller $k$ might suffice for a smaller template. Furthermore, $k$ can also be modulated based on knowledge of the image content. For example, a template with less \textit{texture} content would require a smaller codebook to achieve similar performance. 

\textit{Filter Selection and Granularity.} As seen in Fig. \ref{fig:tlp_ablation} (first and third columns) and Fig. \ref{fig:bbs_ablation} (top two rows), adding simple Haar filters and multi-scale aggregation shows clear gains in performance, especially when using a smaller codebook. This suggests that adding simple spatial information using the filters increases the distribution matching specificity at lower feature dimensions and codebook sizes. As deep features encode information in a larger spatial context, using simpler filters with bigger codebooks and deep features doesn't lead to significant performance gain compared with the base method. 

\textit{Deep Features.} Due to the powerful representation capability, features from pre-trained deep learning models are attractive for template matching. However, approximate NN search methods scale poorly with the feature dimensions \citep{aumuller2020ann} and the template sizes (more computations per pixel). Both speed and recall of the methods are affected negatively by increasing the number of data points or the feature dimensions. Our method shows fractional differences in runtime as seen in Figure \ref{fig:tlp_ablation} (and Tab. \ref{tab:bbs_final}, \ref{tab:tlp_final}) when using deep vs. colour features. Thus, high-dimensional features can be used
to improve the quality of NN matches without adding a significant computational burden. Furthermore, we show in Tab.~\ref{tab:pca} that the runtime and the memory footprint of our methods can be further reduced by dimensionality reduction without much loss in performance, making it suitable for real-time applications. 

Overall, we see the multiple combinations of $K$, $S$, and $\mathcal{F}$'s were able to reach and surpass the DDIS and DIWU performance, showing that our method can achieve high performance while maintaining flexibility. We present and discuss the results of the rotations experiments on BBS25, BBS50, BBS100, and TinyTLP datasets in Supplementary Sec.~\ref{supp:rot_exp}. We show that the performance of our method doesn't degrade as sharply as other NNF-based methods as the query image is rotated.

\section{Conclusion}

We present a fast and robust template-matching framework using vector quantized NNFs, outperforming the previous state-of-the-art methods on high-resolution images while greatly reducing the NN computation cost. Our method efficiently uses the NN matching paradigm, is easily parallelizable to be implemented on a GPU, and scales well with the image size. 

We show a significant reduction in NN computation time can be achieved by using the vector quantization formulation and introducing filtering in the NNFs. Reduction in the NN computations makes using high-dimensional features cost-efficient, leading to better NN matching performance and improving the NNF quality. This paves the way for using deep features efficiently in template-matching tasks. Domain-specific models can be used to boost performance, and different models can be tested quickly using our method.    

The codebook presents a flexible and efficient way of representing the template. In our experiments, each pattern in the codebook was equally weighted; however, the patterns can be prioritized based on human input or prior information. In human-in-the-loop applications, users can click on the patterns to prioritize when the NNF is displayed to the user. Also, patterns can be weighted based on the rest of the template image content, for example, by assigning a higher weight to rarer patterns. Furthermore, the similarity measure is comparable across images, unlike the other fast NNF-based approaches like DIWU and GAD, where the scores are based on image statistics. This can be useful in speeding up the process in applications of template matching where multiple images are considered, e.g., annotation applications and object tracking.

Introducing filtering also presents a flexible and adjustable way of modelling the spatial information in the NNFs. We used Haar filter responses to encode and compare the orientation information between the template and the query image. Complex filters can be designed to improve the performance and the reliability of the algorithm. For example, in industrial automation applications, specific shape-based filters can be designed based on object knowledge for a task to get high performance.

The filters can also be used to extract the orientation information about the detection by comparing the responses of the rotated filters. Rotated filters can be used to extract the orientation information from the detection. For example, multiple rotated Haar filter responses from the query images can be used to approximate the orientation of the detection. This would benefit precision alignment tasks where the template match can be refined based on the rotated filter responses.

The aforementioned advantages make this approach suitable for human-in-the-loop systems where the parameters can be adjusted, and the system's performance can be validated quickly by humans and applied to a large set of images.

\section{Acknowledgement}
The work was supported by the Swedish Foundation for Strategic Research, Grant BD15-0008SB16-0046 and the European Research Council, Grant ERC-2015-CoG 683810.

\bibliographystyle{elsarticle-harv} 
\bibliography{mybib}

\newpage
\section{Supplementary Material}

\subsection{NNF creation}\label{supp:nnf_create}

\begin{algorithm}[h!]
    \algsetup{linenosize=\small}
    \SetAlgoLined{}
    \KwResult{NNF image $N_I$ of size $(h,w)$}
    Codebook, $K$ consisting of $k$ vectors of length $c$\;
    Input image $I$, of size $(h,w,c)$\;
    \For(){i in $h$}{
        \For(){j in $w$}{
            $N_I(i, j)=$ nearest neighbour index of $I(i, j)$ in $K$
        }
    }
    \caption{Creating NNF of the image from a codebook} 
    \label{algo:nnf_codebook}
\end{algorithm}

\subsection{Integral Image}\label{supp:int_img}
The integral image $\iota\iota$ at each location $(x, y)$ of the $k$ dimensional one-hot image $\iota:\mathbb{Z}^2\rightarrow \{0,1\}^k$ is defined as:
\begin{equation}
    \iota\iota(x, y) = \sum_{x'\leq x, y'\leq y} \iota(x', y').
\end{equation}

In an integral image $\iota\iota$, the sum of the values in a rectangular region $A$ of size $(w,h)$ defined by a top-left point $(x_1, y_1)$, top-right point $(x_2, y_1)$, bottom-left point $(x_1, y_2)$, and bottom-right point $(x_2, y_2)$ 
is calculated as:
\begin{equation}
    sum(A) = \iota\iota(x_2, y_2) + \iota\iota(x_1, y_1) - \iota\iota(x_1, y_2) - \iota\iota(x_2, y_1) 
\end{equation}

Hence, the number of calculations for each region now requires only four instead of $wh$, greatly reducing the computational time.

\begin{figure}[tb]
\centering
\includegraphics[width=1\columnwidth]{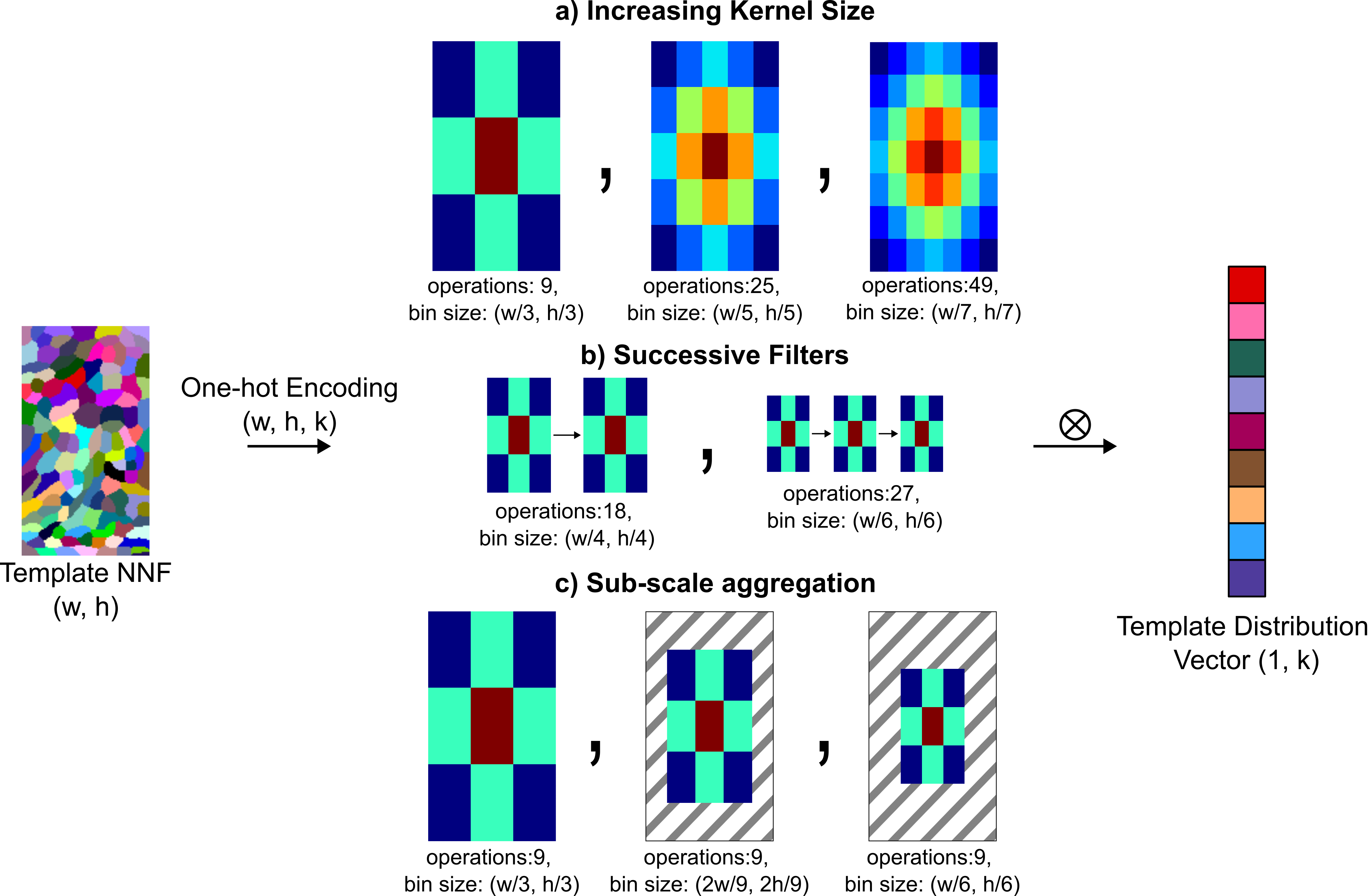}
\caption{Different ways of modulating the bin size for filter response by a) increasing kernel size, b) successive filtering and c) sub-scale aggregation. The grey-shaded region in c) shows the region not considered in the filtering.}
\label{fig:filter_fig}
\end{figure}

\subsection{Modulating Filter Response Granularity} \label{supp:filter_gm}
The granularity of the filter response for a fixed receptive field can be modulated in the following ways (as also illustrated in Fig. \ref{fig:filter_fig}):
\begin{enumerate}
    \item Increasing the filter kernel size. Since the bin size is inversely proportional to the kernel size for a fixed receptive field, increasing the kernel size increases the granularity of the filtering. However, the computations required for each window increase quadratically with the kernel size, which might be undesirable. 
    \item Successive filtering. Applying the filters multiple times increases the granularity exponentially with the dilation rate. At the same time, the number of computations grows linearly.
    \item Sub-scale aggregation. Reducing the receptive field results in finer filter bins and, thus, higher granularity. This enables variable deformation penalties for selective template sub-regions. For example, a higher deformation penalty can be enforced in a template subregion by reducing the receptive field without increasing the number of computations. 
\end{enumerate}

\subsection{Coarse Filter Weight Modification for use on Integral Images}\label{supp:filter_wm}

\begin{algorithm}[tb]
    \algsetup{linenosize=\small}
    \SetAlgoLined{}
    \KwResult{Dilation filter for integral image, $f_d$ of size $(r_w+1, r_h+1)$}
    Original filter, $f_i$ of size $(r_w, r_w)$\;
    Integral Multiplier, $m=[[1, -1 ], [-1, 1]]$\;
    Initialize \(f_d\) with zeros\;
    \For(){i in $r_w$}{
        \For(){j in $r_h$}{
            $f_d(i:i+1, j:j+1) \mathrel{{+}{=}} f_i(i, j)*m$
        }
    }
    \caption{Modifying dilation filter weights for region aggregation on the integral image} 
    \label{algo:conv_filt}
\end{algorithm}

Let $f_i$ be a coarse filter of kernel size $(r_w, r_h)$ with each weight repeated over the rectangular bins of size $(d_x, d_y)$. The dilation filter $f_d$ would then have the kernel size of $(r_w+1, r_h+1)$. The filter weights of the dilation filter are calculated using Algorithm \ref{algo:conv_filt}. Thus, the number of calculations per window has been reduced from $r_w\cdot d_x\cdot r_h\cdot d_y$ to $(r_w+1)\cdot(r_h+1)$ making the filtering process efficient.

\subsection{Rotation Experiments} \label{supp:rot_exp}

\begin{figure}[tb]
\centering
\includegraphics[width=1\columnwidth]{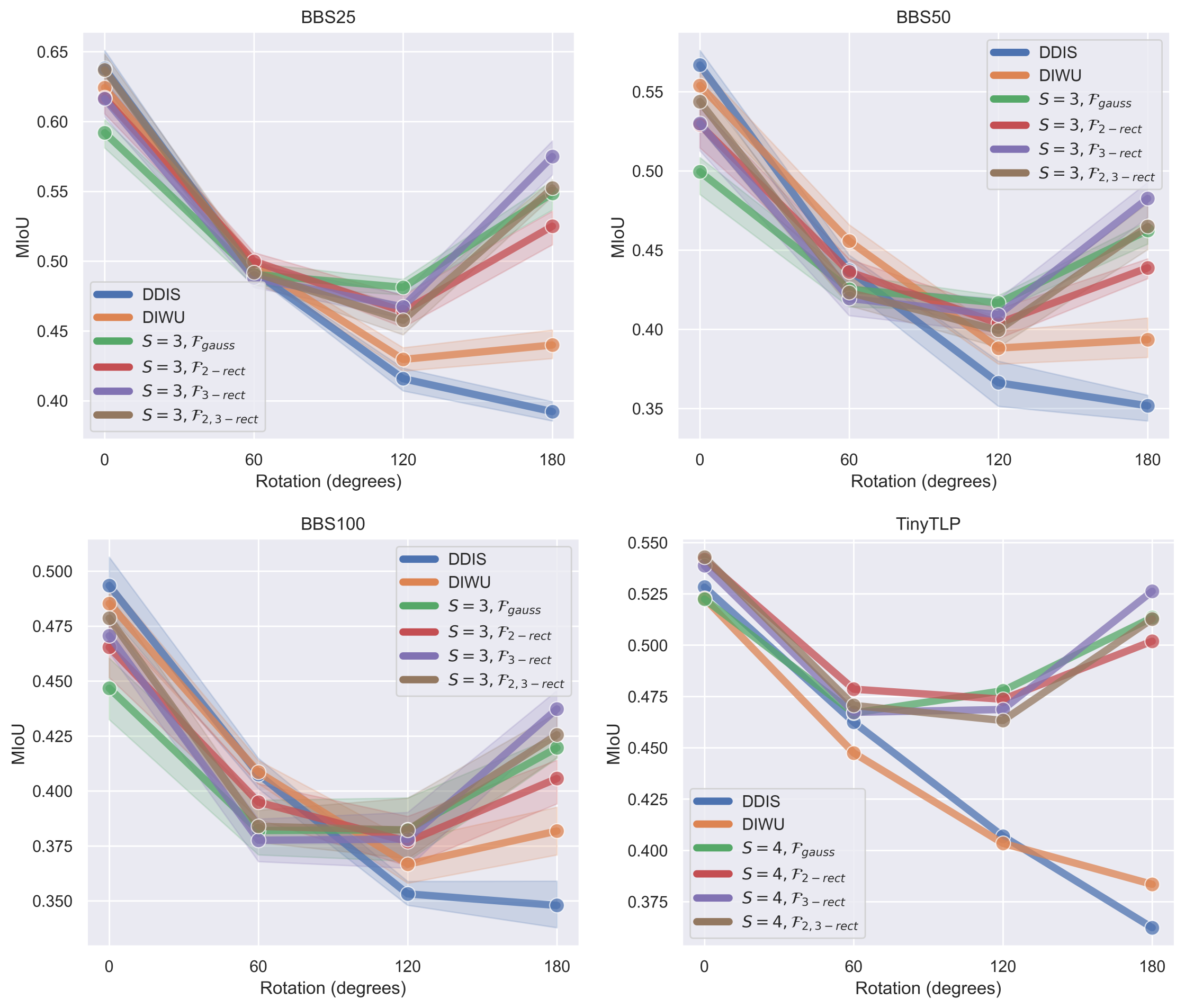}
\caption{MIoU performance of DDIS, DIWU, and our method for the rotation experiments on the BBS25, BBS50, BBS100, and TinyTLP datasets.}
\label{fig:rot_comp}
\end{figure}

We constructed new datasets from BBS and TinyTLP by rotating the query images in the dataset by an angle $\theta$ while keeping the same template image and box. The query box is also rotated at the same angle and treated as the new ground truth. The prediction box dimensions are chosen to be the same as the rotated template box dimensions for better comparison. We considered $\theta \in \{60\degree,120\degree,180\degree\}$ and compared the results with the original dataset performance without rotation. We chose to show the performance of our method on color features and highlight the results of all the spatial filters on our best-performing aggregation scale. Deep features were not considered as the runtime of DDIS and DIWU with deep features is quite high.

Fig. \ref{fig:rot_comp} shows the performance of the rotation experiment on the datasets. DDIS and DIWU performance drops sharply with increased rotation. As expected, DDIS and DIWU are not robust against large rotations as their deformation penalty relies on the pixel distance. However, the performances of the different configurations of our method do not degrade as sharply. For BBS datasets, the performance of our method at $\theta=60\degree$ is close to that of DDIS and DIWU; however, our method outperforms them at higher rotational deformations, $\theta=120\degree,180\degree$. For the TinyTLP dataset, our method outperforms DDIS and DIWU for every rotation angle. DDIS and DIWU perform well in the original dataset, however, with a higher penalty for larger deformations, DDIS and DIWU are not robust against large rotations. Our methods perform similarly to DDIS and DIWU for the original dataset, but the performance doesn't degrade as severely when rotations are present. The performance of our methods increases back up at $\theta=180\degree$, which can be attributed to the filters used. The similarity score from the Gaussian filter representations would be high for $\theta=180\degree$ as the filters are symmetric around the center.

\subsection{Design Choices Evaluation}\label{supp:bbs_dce}

\newpage
\subsection{BBS dataset Qualitative results}\label{supp:bbs_vis}

\begin{figure}[!tb]
\centering
\includegraphics[width=1\columnwidth]{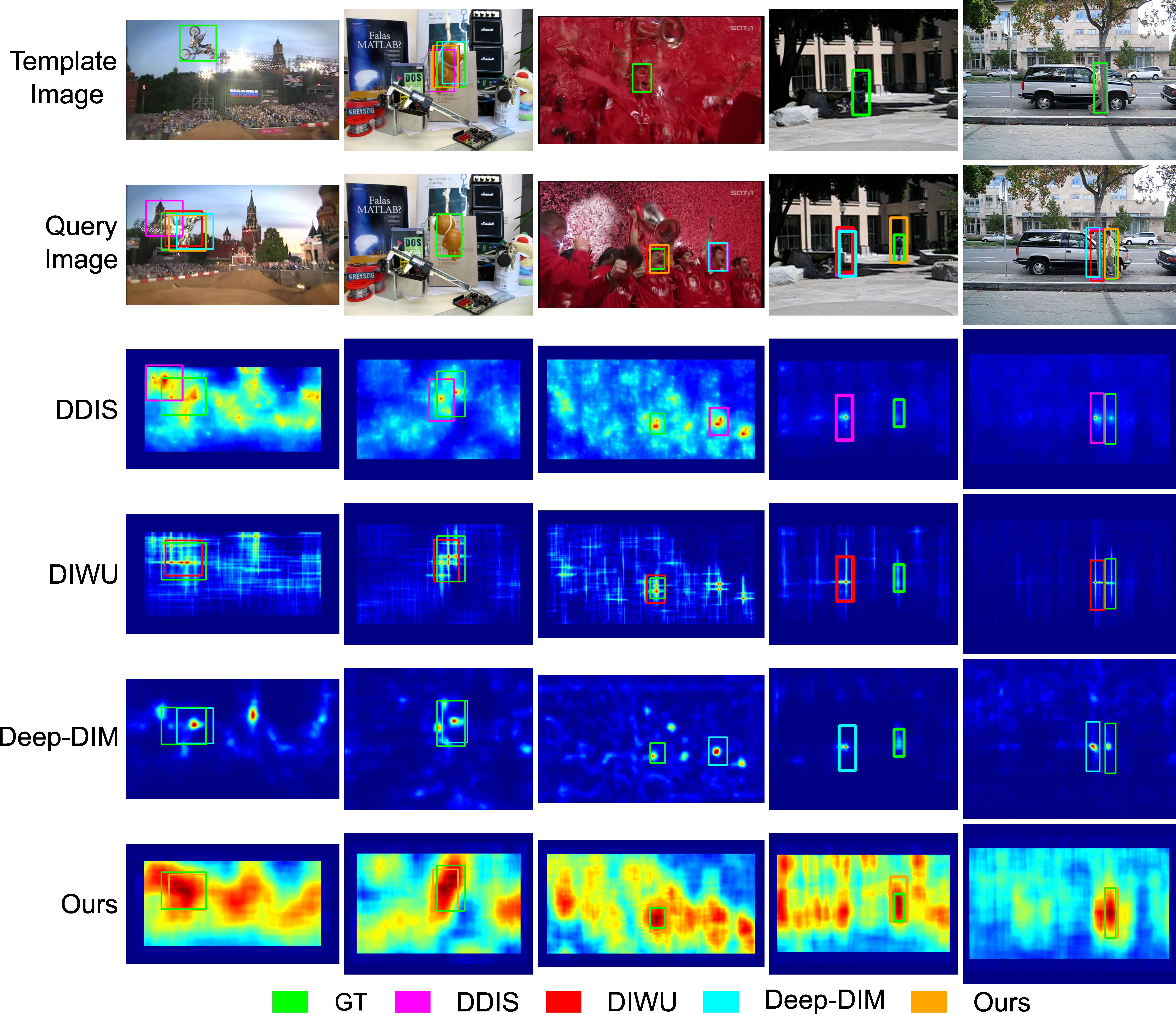}
\caption{Qualitative results for the BBS dataset using deep features. Images were chosen to highlight the differences between the methods. The first column shows the template image with the template marked in green. The second column shows the query image with the results of DDIS (in pink), DIWU (in red), Deep-DIM (in blue) and our method (in orange). The third to sixth columns show the heatmaps of the DDIS, DIWU, Deep-DIM, and our method, respectively, with the predicted bounding box marked in the same colours as the second column.}
\label{fig:bbs_vis_comp}
\end{figure}

\end{document}